\documentclass[12pt]{l4dc2021} 

\title[Safe RL With RAG]{Safe Reinforcement Learning Using Robust Action Governor}
\usepackage{times} 
\usepackage{optidef}
\usepackage{algorithm} 
\usepackage{algpseudocode} 
\usepackage{romannum}
\DeclareMathOperator*{\argmin}{arg\,min}
\DeclareMathOperator*{\argmax}{arg\,max}

\author{%
 \Name{Yutong Li} \Email{yutli@umich.edu}\\
 \addr University of Michigan, Ann Arbor, MI, USA
 \AND
 \Name{Nan Li} \Email{nanli@umich.edu}\\
 \addr University of Michigan, Ann Arbor, MI, USA
 \AND
 \Name{H. Eric Tseng} \Email{htseng@ford.com}\\
 \addr Ford Motor Company, Dearborn, MI, USA
 \AND
 \Name{Anouck Girard} \Email{anouck@umich.edu}\\
 \addr University of Michigan, Ann Arbor, MI, USA
 \AND
 \Name{Dimitar Filev} \Email{dfilev@@ford.com}\\
 \addr Ford Motor Company, Dearborn, MI, USA
 \AND
 \Name{Ilya Kolmanovsky} \Email{ilya@umich.edu}\\
 \addr University of Michigan, Ann Arbor, MI, USA
}

\begin{document}

\maketitle

\begin{abstract}%
  Reinforcement Learning (RL) is essentially a trial-and-error learning procedure which may cause unsafe behavior during the exploration-and-exploitation process. This hinders the application of RL to real-world control problems, especially to those for safety-critical systems. In this paper, we introduce a framework for safe RL that is based on integration of a RL algorithm with an add-on safety supervision module, called the Robust Action Governor (RAG), which exploits set-theoretic techniques and online optimization to manage safety-related requirements during learning. We illustrate this proposed safe RL framework
  through an application to automotive adaptive cruise control.
\end{abstract}


\section{Introduction}\label{sec:1}
In Reinforcement Learning (RL), the agent interacts with the environment by perceiving environment's states and selecting the action that maximizes the long-term return based on a real-valued reward signal \citep{sutton2018}. The success of RL has been apparent in a broad spectrum of applications \citep{mnih2013,kober2013,abbeel2007}. However, RL is essentially a trial-and-error learning process which may cause unsafe behavior during the learning process. This hinders the RL real-world applications, especially to safety-critical systems.

One approach to addressing such safety issues is the safe RL approach \citep{garcia2015}. In particular, risk-sensitive safe RL aims to promote the constraint satisfaction via balancing the long-term return and the risk of reaching the unsafe region \citep{geibel2005}. Methods based on policy optimization with constraints were also proposed, where a constraint on the probability of the system being able to return to the safe region is enforced \citep{wachi2018,moldovan2012}. However, safety cannot be guaranteed via these model-free approaches, as the agent needs to learn to operate safely via its interactions with the environment, which may lead to constraint violations. 

To guarantee the safety constraint satisfaction during the learning process, an effective approach is to incorporate the system model information. In general, the safe region can be determined by the system model, and a control policy that keeps system state staying inside this safe region can then be computed \citep{aswani2013,fisac2018,larsen2017,sloth2012}. The main advantage of these model-based methods is that within the interior of the safe region, the RL agent can explore safely to improve the performance. However, the optimality of the trained policy is highly dependent on the size of the safe region computed based on the system model. 

Following the idea of this model-based direction, in this paper, we propose a novel safe RL framework that also exploits the system model to design a safe set that regulates the RL explorations to guarantee system safety. In particular, the proposed framework exploits an add-on module, called the Robust Action Governor (RAG), to manage safety. With the RAG, an arbitrary RL algorithm can be integrated into the framework and lead to safe RL. The RAG enforces safety constraints by monitoring, and minimally modifying when necessary, the control signal produced by the nominal RL policy to a constraint-admissible one. 

The proposed safe RL framework based on RAG operates on the basis of set-theoretic techniques and online optimization. Similar approaches were proposed by integrating the Reference Governor (RG) with RL to enforce constraint satisfaction \citep{li2018,li2019}. A distinguishing feature of our safe RL framework based on RAG is that, unlike the RG which modifies the reference input to the controller, the RAG modifies controller output signal, i.e. the RAG can be placed closer to the environment/plant. A direct consequence is that our safe RL framework based on RAG can be used to train lower-level controls, while the approach based on RG in \cite{li2018,li2019} can only be used to train higher-level planners where the plant must have already been stabilized by some controller that is fixed. Another advantage of RAG compared to RG is that the use of RAG yields a larger safe set, and thereby, can potentially achieve better control performance \citep{9109265}. On the other hand, unlike safe RL approaches based on the use of control barrier functions \citep{cheng2019end,sloth2012}, our RAG is formulated based on discrete-time models from the start and thereby can be more directly applied in a digital setting (e.g., one does not need to modify the algorithm to account for sampling time and discrete updates). 

In summary, the contributions of this paper include: 1) establishing a safe RL framework based on RAG, 2) extending the theoretical results and computational methods in \cite{9109265} for the Action Governor for discrete-time linear systems to their robust versions for linear systems with additive bounded disturbances and non-convex constraints, and 3) illustrating effectiveness of the proposed safe RL framework using an example relevant to automated driving and adaptive cruise control.


\section{Conventional RL and Safe RL}\label{sec:2}
\begin{figure}[thpb]
      \centering
      \includegraphics[scale=0.5]{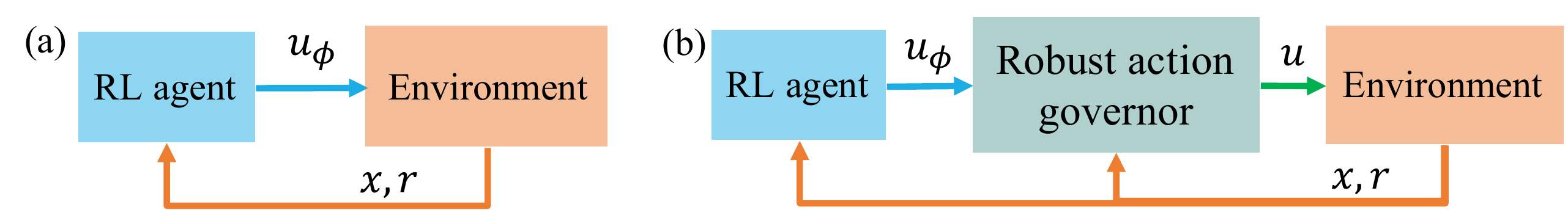}
      \caption{Schematic diagram of conventional RL (a) and safe RL (b).}
      \label{fig:ConceptFig}
\end{figure}

Conventional RL optimizes the agent's action via trial-and-error to maximize its accumulated reward (or minimize accumulated cost) through continuously interacting with the environment, as illustrated in Figure \ref{fig:ConceptFig}a. Specifically, at
each time instant $k \in \mathbb{Z}_{\geq 0}$, the agent takes a measurement of the state $x(k)\in \mathbb{R}^n$, executes a control $u_\phi(k)\in \mathbb{R}^m$ and collects a reward $r(k)\in \mathbb{R}$. A control policy, $\pi: \mathbb{R}^n \rightarrow \mathbb{R}^m$, which is a mapping from the state space $\mathbb{R}^n$ to the action space $\mathbb{R}^m$, describes the agent’s behavior. The control policy is learned from the experience $\{x(k),u_\phi(k),r(k)\}$ to maximize the long-term reward $R=\sum_{k=0}^{\infty} \gamma^k r(k)$, where $\gamma \in (0,1)$ is a discount factor. However, to maximize the long-term reward, RL agent needs to explore within the action space, which may cause unsafe behaviors (i.e., violation of certain safety constraints) during the learning process. This feature hinders the application of RL to actual engineering systems, and motivates us to propose a safe RL framework enabling the agent to learn a policy safely.  

The proposed safe RL framework is illustrated in Figure \ref{fig:ConceptFig}b. An add-on module, termed Robust Action Governor (RAG), is introduced between the RL agent and the environment (which represents the system that the control policy is acting on, e.g., the plant in a conventional control setting). The RAG monitors the control signal $u_\phi(k)$ generated by the RL agent and corrects the ones that may cause unsafe behavior. We will describe the properties and design procedure of the RAG in the subsequent section. \\

\section{Robust Action Governor}\label{sec:3}
\subsection{Problem Formulation}\label{sec:31}
In this section, we review the main results in \cite{9109265} on the Action Governor and extend them from disturbance-free systems to systems subject to bounded disturbances, enabling the Action Governor to be applicable to more general cases. Consider a discrete-time linear model with additive disturbances as follows:
\begin{equation} \label{equ:SysDyn}
   x(k+1)=Ax(k)+Bu(k)+Ew(k),
\end{equation}
where $x(k)\in \mathbb{R}^n$ is the state at discrete time instant $k\in \mathbb{Z}_{\geq 0}$ and $u(k)\in \mathbb{R}^m$ is the control input. The vector $w(k)$ represents an unmeasured disturbance input and we assume that it is bounded in a known set, i.e., $w(k) \in \mathcal{W} = \{ w:Mw \leq m\}$, where $\mathcal{W}$ is a known polytope in $\mathbb{R}^{n_w}$. In applications to RL, $w(k)$ can account for model mismatch, e.g., differences between the actual lead vehicle acceleration and assumed zero lead vehicle acceleration in an adaptive cruise control (ACC) setting. We further assume that state, input and disturbance constraints are not coupled, i.e., $\mathcal{T}=\{(x,u,w)|x\in \mathcal{X}, u\in \mathcal{U}, w\in \mathcal{W}\}=\mathcal{X}\times \mathcal{U}\times \mathcal{W}$, where $\mathcal{X}$ represents the feasible set of states and $\mathcal{U}$ is the control input set. We assume that a nominal control policy $\phi$ has been designed for the system (\ref{equ:SysDyn}), 
\begin{equation} \label{equ:nominalCtr}
   u_{\phi} = \phi(x(k),x_r(k),w(k),k),
\end{equation}
where $x_r(k) \in \mathbb{R}^r$ denotes a reference signal for the system \eqref{equ:SysDyn}, which defines the control objective. Note that there is no further assumption on the nominal control $\phi$, and it can take any form, e.g. nonlinear and time-varying. In particular, in this paper, $\phi$ represents the RL policy.

We assume that the system is subject to a safety requirement of the form 
\begin{equation} \label{equ:ExcluZone}
   x(k) \in \mathcal{X} = \mathbb{R}^n \setminus X_0,\ \ \  \forall k \in \mathbb{Z}_{\geq 0},
\end{equation}
where the unsafe set $X_0$ can be expressed as a finite union of polytopes, i.e., 
\begin{equation} \label{equ:X0_Definition}
   X_0 = \bigcup\limits_{i=1}^{n_g} \{x\in \mathbb{R}^n:G_{i}x<g_{i}\},
\end{equation}
where $G_i \in \mathbb{R}^{n_g \times n}$ and $g_i \in \mathbb{R}^{n_g}$. Note that the expressions \eqref{equ:ExcluZone} and \eqref{equ:X0_Definition} can represent a broad range of safety requirements, including box constraints, obstacle avoidance constraints, etc. In general, the feasible set $\mathcal{X}$ can be non-convex.

As the safety requirement (\ref{equ:ExcluZone}) may not be strictly handled by the nominal control policy \eqref{equ:nominalCtr} (especially when \eqref{equ:nominalCtr} represents an RL policy), this motivates us to design an add-on scheme to enforce (\ref{equ:ExcluZone}). In particular, we exploit a supervisory solution, the RAG, to satisfy this requirement, as illustrated in Figure \ref{fig:ConceptFig}b. The RAG monitors the nominal control input $u_{\phi}$ and, if necessary, minimally modifies $u_{\phi}$ to guarantee that the system state can stay outside $X_0$ for the present and all future time instants even in the presence of disturbances. 

In particular, at each time instant, the RAG solves the following constrained optimization problem to enforce the safety requirement (\ref{equ:ExcluZone}), 
\begin{subequations}\label{equ:RAG}
\begin{align}
    u(k) =& \argmin_{u \in \mathcal{U}}\ {\|u-u_{\phi}(k)\|}_S^2 \label{equ:RAG1}\\
    & \text{ subject to }\, Ax(k)+Bu+Ew \in X_{\text{safe}},\forall w \in \mathcal{W}\label{equ:RAG2}
\end{align}
\end{subequations}
where $X_{\text{safe}}\subset \mathcal{X}$ is a ``safe set'' which will be introduced in the next section. The function $\| \cdot \|_S=\sqrt{(\cdot)^TS(\cdot)}$ is used to penalize the difference between the nominal control $u_{\phi}$ and the modified control $u$, where the matrix $S\in \mathbb{R}^{m\times m}$ is positive-definite.


\subsection{Safe Set and Unrecoverable Sets}\label{sec:32}
To enforce both present and future safety, the safe set $X_{\text{safe}}$ is characterized by the following requirements: There exists a state-feedback control $u(x)$ such that
\begin{itemize}
  \item $u(x) \in \mathcal{U}$ for all $x \in X_{\text{safe}}$;
  \item Given $x(0) \in X_{\text{safe}}$, this control can keep all future states $\{x(1),x(2),...\}$ within $\mathcal{X}$ in the presence of disturbances $\{w(0),w(1),...\} \subset \mathcal{W}$.
\end{itemize}


The derivation of $X_{\text{safe}}$ relies on its complementary sets, termed unrecoverable sets, which are recursively defined as follows,
\begin{align} \label{equ:UnrecoverSetLinear}
\begin{split}
    X_k & = X_0 \cup \{ x \in \mathbb{R}^n: \forall u \in \mathcal{U}, \exists w \in \mathcal{W},\text{s.t.}\  Ax+Bu+Ew \in X_j\\
     & \quad \text{for some}\, j=0,...,k-1\}\\
     & = X_0 \cup \left\{ x \in \mathbb{R}^n: Ax\in \bigcup\limits_{j=0}^{k-1} X_{j} \oplus ((-E) \circ\mathcal{W}) \sim (B \circ \mathcal{U})\right\},
\end{split}
\end{align}
where $\oplus$, $\sim$, and $\circ$ represent the Minkowski sum, Pontryagin difference, and affine mapping operations of sets. 
Note that in the derivation of (\ref{equ:UnrecoverSetLinear}) we have used the fact that $S=\{v:\forall u\in U, \exists z\in Z,\text{s.t.}\ v=z-u\} =\{v:\forall u\in U, \exists z\in Z,\text{s.t.}\ z=v+u\}=\{v:v+u\in Z,\forall u\in U\}=Z\sim U$.

The unrecoverable set $X_k$ in  (\ref{equ:UnrecoverSetLinear}) has the following properties elaborated in Propositions 1 to 3.

\noindent \textbf{Proposition 1:}
1) If $x_0 \in X_k$, then for any state-feedback control sequence $\{u_0(x_0)...,u_{k-1}(x_{k-1})\\\} \in \mathcal{U} \times ...\times \mathcal{U}$, there exists a disturbance sequence $\{w_0,...,w_{k-1}\}\in \mathcal{W} \times ...\times \mathcal{W}$ such that $x_j \in X_0$ for some $0 \leq j \leq k$ .
\\
2) Let $x_0$ be given. If for any state-feedback control sequence $\{u_0(x_0),...,u_{k-1}(x_{k-1})\}\in \mathcal{U} \times ...\times \mathcal{U}$, there exists a sequence $\{w_0,...,w_{k-1}\}\in \mathcal{W} \times ...\times \mathcal{W}$ such that $x_j \in X_0$ for some $0 \leq j \leq k$, then $x_0 \in X_k$.

The proof of Proposition 1 can be constructed following similar steps as those in the proofs of Propositions 1 and 2 in \cite{9109265}. It is omitted here but will be included in our subsequent publication \cite{RAGHybrid}. Proposition 1 implies that for any state-feedback control sequence $\{u_0(x),...,u_{k-1}(x)\}\in \mathcal{U}\times...\times \mathcal{U}$, there exists a disturbance sequence $\{w_0,...,w_{k-1}\}\in \mathcal{W}\times...\times \mathcal{W}$ that causes the state trajectory to enter $X_0$ 
during the steps $0,...,k$ if and only if $x_0 \in X_k$. 
Or equivalently, for any admissible disturbances $\{w_0,...,w_{k-1}\}\in \mathcal{W}\times...\times \mathcal{W}$, there exists a state-feedback control sequence $\{u_0(x),...,u_{k-1}(x)\}\in \mathcal{U}\times...\times \mathcal{U}$ that can prevent the state trajectory from entering $X_0$ over steps $0,...k$ if and only if $x_0 \in \mathcal{X} \setminus X_k$.



\noindent \textbf{Proposition 2:} 
For each $k=0,1,...,$ we have $X_k \subset X_{k+1}$, i.e. $X_k$ is an increasing sequence of sets. In turn, $X_{\infty}=\lim_{k\to\infty} X_k$ exists and satisfies $X_k \subset X_{\infty}$ for all $k$.

On the basis of the unrecoverable sets $X_k$, we define the safe set as $X_{\text{safe}}=\mathcal{X} \setminus X_{\infty}=\lim_{k\to\infty} (\mathcal{X} \setminus X_k)$.

\noindent \textbf{Proposition 3:}
For any $x \in X_{\text{safe}}$, it holds that (\romannum{1}) $x \in \mathcal{X} = \mathbb{R}^n \setminus X_0$ and (\romannum{2}) there exists $u \in \mathcal{U}$ such that $Ax + Bu + Ew \in X_{\text{safe}}$ for all $w\in \mathcal{W}$.

The proofs of the above Propositions 2 and 3 can also be constructed following proofs of the Propositions 3 and 4 in \cite{9109265} and will be included in \cite{RAGHybrid}. Note that one needs to use the formula $\bigcup\limits_{k=0}^{k} X_k \oplus W = \bigcup\limits_{k=0}^{k} (X_k \oplus W)$ to complete these proofs, which can be easily shown to hold using the definitions of Minkowski sum and union of sets.

Proposition 3 ensures that if the RAG operates based on (\ref{equ:RAG}), then a feasible solution exists to (\ref{equ:RAG}) for all $k$, and the safety requirement (\ref{equ:ExcluZone}) is satisfied for all $k$ in spite of disturbance $w\in \mathcal{W}$.
Note that the exact determination of $X_{\text{safe}}$ relies on the set $X_k$ iteratively computed according to (\ref{equ:UnrecoverSetLinear}) with $k\rightarrow \infty$.
In practice, we approximate $X_{\text{safe}}$ by $X_{\text{safe},k} = \mathcal{X}\setminus X_{k}$ with $k$ being sufficiently large. 
We also note that according to Proposition 1, the safe set $X_{\text{safe}}$ used by RAG approach is maximal.


\subsection{Offline and Online Computations}\label{sec:33}

\noindent \textbf{Proposition 4:}
Suppose $A$ is invertible. Then, for each $k=1,2,...,$ we have (\romannum{1}) $X_k$ can be represented as the union of a finite number of polytopic sets, i.e., $X_k=\bigcup\limits_{j=1}^{r_k} X_{k,j}$ where $X_{k,j}$ is a polytopic set for each $j=1,...,r_k$; and (\romannum{2}) $X_k$ can be numerically computed using Algorithm \ref{AlgoUnrecoverSet}.

The proof of Proposition 4 is similar to that of Proposition 6 in \cite{9109265}. RAG operates by solving the optimization problem in (\ref{equ:RAG}) at each time step. Using the fact that $\mathcal{X} \backslash (X_{k} \oplus (-E\circ\mathcal{W}))=\mathcal{X} \backslash X_{k} \sim E\circ\mathcal{W}$, and $X_{k}\oplus (-E\circ\mathcal{W}) = \bigcup\limits_{j=1}^{r_{k}} \bigcap \limits_{i=1}^{s_{j}}\{x\in \mathcal{X}:G_{i,j}x<g_{i,j}\}$, (\ref{equ:RAG}) can be transformed into a Mixed-Integer Quadratic Programming (MIQP) problem with the constraints as following
\begin{subequations}\label{equ:RAGMIQP}
\begin{align}
    G_{i,j}(Ax(k)+Bu)\geq g_{i,j}-M(1-\delta_{i,j}), \label{equ:RAGMIQP1}\\
     \delta_{i,j} \in \{0,1\},\forall i=1,...,s_j,\forall j=1,...,r_{k^{'}},\label{equ:RAGMIQP2}\\
     \sum\limits_{i=1}^{s_j} \delta_{i,j} =1, \forall j=1,...,r_{k^{'}},\label{equ:RAGMIQP3}
\end{align}
\end{subequations}
where $M>0$ is a sufficiently large number. 

\begin{algorithm}
        \caption{Offline computation of $X_k$ for systems in (\ref{equ:SysDyn})}
        \hspace*{\algorithmicindent} \textbf{Input:} {$A,B,E,X_0,X_{k-1},\mathcal{U},\mathcal{W}$}\\
        \hspace*{\algorithmicindent} \textbf{Output:}{$\ X_k$}
        \label{AlgoUnrecoverSet}
        \begin{algorithmic}[1] 
            \State $\mathcal{H} \gets \text{convhull}(X_{k-1}\oplus ((-E) \circ \mathcal{W}))$
            \State $\mathcal{D} \gets \mathcal{H} \sim ((B) \circ \mathcal{U}))$
            \State $\mathcal{E} \gets \mathcal{H} \setminus  (X_{k-1}\oplus ((-E) \circ \mathcal{W})))$
            \State $\mathcal{F} \gets \mathcal{E} \oplus  ((-B) \circ \mathcal{U})$
            \State $\mathcal{G} \gets \mathcal{D} \setminus  \mathcal{F}$
            \State $X_{k} \gets X_0 \bigcup A^{-1}\mathcal{G}$
        \end{algorithmic}
\end{algorithm}

\section{Safe RL With RAG}\label{sec:4}
Given the constraint enforcement property of RAG, in this section, we integrate RL with RAG to achieve safe learning without violating system constraints. Indeed, the proposed safe RL framework can be combined with any RL algorithms. In this paper, we consider the Neural-Fitted Q-learning (NFQ) as the base RL module due to its ability to deal with continuous state and control spaces \citep{riedmiller2005neural}.

In NFQ, we use a neural network to approximate the Q-function, and we update this Q-function approximation using the following formula,
\begin{equation}\label{equ:Q_learning}
 Q(x(t),u(t))\leftarrow \lambda \tilde{Q}(x(t),u(t)) + (1-\lambda) \big(R(x(t),u(t)) + \gamma \tilde{V}(x(t+1))\big),
\end{equation}
where $\lambda \in (0,1)$ is the learning rate, $\gamma \in (0,1)$ is a factor to discount future rewards and avoid the Q-values increasing to infinity, and
\begin{equation}\label{equ:Q_learning_V}
 \tilde{V}(x) = \max_{u \in \mathcal{U}} \tilde{Q}(x,u),
\end{equation}
where $\tilde{Q}(x,u)$ represents the approximated Q-values extracted from the neural network. Note that the data points used to update Q-function in (\ref{equ:Q_learning}) can be obtained through black-box simulations or hardware experiments, which makes the NFQ a model-free method.

The proposed safe RL framework based on RAG is formally presented as Algorithm \ref{RL_RAG}. We first initialize a neural network to approximate Q-function for the continuous state and action spaces. The action space $\mathcal{U}$ is discretized with step $d_s$ to reduce the computational cost of selecting the optimal control action $u(t)$ based on currently approximated Q-function in (\ref{equ:Q_learning_V}). The $n_t$ is the collected trajectory number in one episode, and $T$ is the length of one trajectory. We use a replay buffer $B$ to store the experience collected in the most recent episode.



The RL agent balances exploration and exploitation using an $\epsilon$-greedy action selection rule in Lines~5-9. Line 10 is the control modification step. By solving (\ref{equ:RAG}), any action that may lead to constraint violation will be modified to $u^{\text{safe}}$. Note that we only discretize the action space $\mathcal{U}$ in Line 8, (\ref{equ:RAG}) is solved on the continuous action space. 


Lines 11-16 collect experience and store it into replay buffer $B$ for later use in training the Q-function NN $\tilde{Q}(x,u)$. In particular, we update the Q-value of the current state $x(t)$ and nominal control $u(t)$, $Q(x(t),u(t))$, with the reward $R(x(t),u^{\text{safe}}(t))$ brought by safe action $u^{\text{safe}}(t)$. Furthermore, we store the tuple of current state, nominal control and updated Q-value, $(x(t),u(t),Q(x(t),\\u(t)))$, to the replay buffer. This way, the action modification by RAG is not perceived by the agent, and the agent will explore all actions. This may be beneficial as this will potentially improve convergence to the optimal control policy.

\begin{algorithm}
	\caption{Safe RL algorithm}
	\hspace*{\algorithmicindent} \textbf{Input} {Initialized Q-value NN $\tilde{Q}(x,u)$, empty replay buffer $B$, discretized action space $\mathcal{U}_d$, and the maximum trajectory number within a training episode $N$.}
	\begin{algorithmic}[1]
		\For {each training episode}
		\While{$n_t < N$}
		\State{Pick a safe initial state $x(0)\in X_{\text{safe}}$}
		    \While{$t<T$}
                \If{$rand()<\epsilon$}
                    \State $u(t)$ takes a random value within $\mathcal{U}_d$\algorithmiccomment{Exploration}
                \Else
                    \State {$u(t)\in\argmax_{u\in \mathcal{U}_d} \tilde{Q}(x(t),u)$}\algorithmiccomment{Exploitation}
                \EndIf
                \State{Solve (\ref{equ:RAG}) to modify $u(t)$ to $u^{\text{safe}}(t)$}\algorithmiccomment{RAG}
                \State{Apply $u^{\text{safe}}(t)$ to system}
                 \State{Observe next state $x(t+1)$ and reward $R(x(t),u^{\text{safe}}(t))$}
                 \State{Update Q value:}
                 \State $\tilde{V}(x(t+1)) = \max_{u\in \mathcal{U}_d}\tilde{Q}(x(t+1),u)$
                 \State{$Q(x(t),u(t))\leftarrow \lambda \tilde{Q}(x(t),u(t)) + (1-\lambda)\big(R(x(t),u^{\text{safe}}(t)) + \gamma \tilde{V}(x(t+1))\big)$}
                \State{Store $(x(t),u(t),Q(x(t),u(t)))$ to replay buffer $B$}
		    \EndWhile
		    \EndWhile
		    \State{Train Q-value NN $\tilde{Q}(x,u)$ using sampled data from replay buffer $B$}
		    \State $n_t \leftarrow 0$
		\EndFor
	\end{algorithmic} 
	\label{RL_RAG}
\end{algorithm} 

Note that the proposed RAG approach can be combined with an arbitrary RL algorithm to achieve safe RL. The discretization step of action space $\mathcal{U}$ is only for reducing the computational complexity and is not a necessary step of the proposed safe RL framework. Furthermore, although in Algorithm \ref{RL_RAG} the Q-learning algorithm produces a nominal control input that takes values in the discretized space $\mathcal{U}_d$, the modified safe control input $u^{\text{safe}}$ after RAG takes values in the original continuous space.
\section{Safe RL for Adaptive Cruise Control}\label{sec:5}
In this section, we apply the proposed safe RL framework to the Adaptive Cruise Control (ACC) problem for an automated vehicle, to demonstrate the major advantage of the proposed safe RL approach over conventional RL in terms of guaranteed constraint satisfaction both during and after training. The dynamics of relative motion between the lead vehicle and the following ego vehicle are represented as
\begin{align}\label{equ:ACC}
\begin{split}
 \begin{bmatrix} \Delta s(k+1) \\ \Delta v(k+1) \\ v^{\text{ego}}(k+1)\end{bmatrix}
 =&
  \begin{bmatrix}
   1 & T_s & 0\\
   0 & 1 & 0\\
   0 & 0 & 1
   \end{bmatrix}
   \begin{bmatrix} \Delta s(k) \\ \Delta v(k) \\ v^{\text{ego}}(k)\end{bmatrix} + 
   \begin{bmatrix} -\frac{T_s^2}{2} \\ -T_s \\ T_s
   \end{bmatrix} u(k) +\begin{bmatrix} \frac{T_s^2}{2} \\ T_s \\ 0 \end{bmatrix}w(k),
\end{split}
\end{align}
where $\Delta s$[m] and $\Delta v$[m/s] are the longitudinal distance and relative speed between the lead and ego vehicles, respectively. The state $v^{\text{ego}}$[m/s] represents the ego vehicle's velocity. The control input $u$[m/s$^2$] represents the ego vehicle's acceleration, and to improve the driving comfort, we impose the following input constraints $u\in \mathcal{U}=\{u:-3 \leq u \leq 3\}$. $T_s=0.5$s is the sampling period. We treat the lead vehicle's acceleration $w\in \mathcal{W}=\{w:-1.5 \leq w \leq 1.5\}$ as a disturbance. The upper and lower bounds of the lead vehicle's acceleration are derived based on the FTP75 driving cycle, which represents the typical city driving behavior.

Our goal is to achieve a safe and efficient car-following performance. We use headway time as the metric and set its target value to 1.5s. The reward function of the RL agent is design as following:
\begin{equation}
\label{equ:ACC_reward}
R
 =\begin{cases}
   -(\frac{\Delta s}{v^{\text{ego}}}-1.5)^2 & \text{if $v^{\text{ego}} \geq 5$},\\
   
  -(\Delta s-7.5)^2 & \text{if $0\leq v^{\text{ego}} \leq 5$}.\\
\end{cases}
\end{equation}
Also, we consider the following constraints on the headway time to enforce safety:
\begin{equation}\label{equ:ACCSpec}
 1 \leq \frac{\Delta s}{\text{max}(v^{\text{ego}},5)} \leq 2.
\end{equation}
As the set defined by \eqref{equ:ACCSpec} is the region where the system state is supposed to be in, $X_0$ is its complement and can be expressed in the form of (\ref{equ:X0_Definition}).

The nominal control $u_\phi$ (before training) is a state-feedback control policy tracking the desired headway time of 2.5s, which is different from the target, i.e. 1.5s. Two RL schemes, namely, conventional RL and safe RL, as shown in Figure \ref{fig:ConceptFig}, are employed to train the nominal control policy $u_\phi$. During training, we randomly sample segments (with length of $T=30$s) within the FTP75 driving cycle as the lead vehicle's speed trajectory. The unrecoverable set $X_k$ and safe set $X_{\text{safe},k}$ are computed with the MPT3 toolbox offline \citep{MPT3}. The online MIQP optimization problem in (\ref{equ:RAG}) is solved by OPTI with SCIP \citep{CW12a,Achterberg2009}. We compute $X_k$ according to Algorithm \ref{AlgoUnrecoverSet} with $k=10$.

\begin{figure}[thpb]
      \centering
      \includegraphics[scale=0.51]{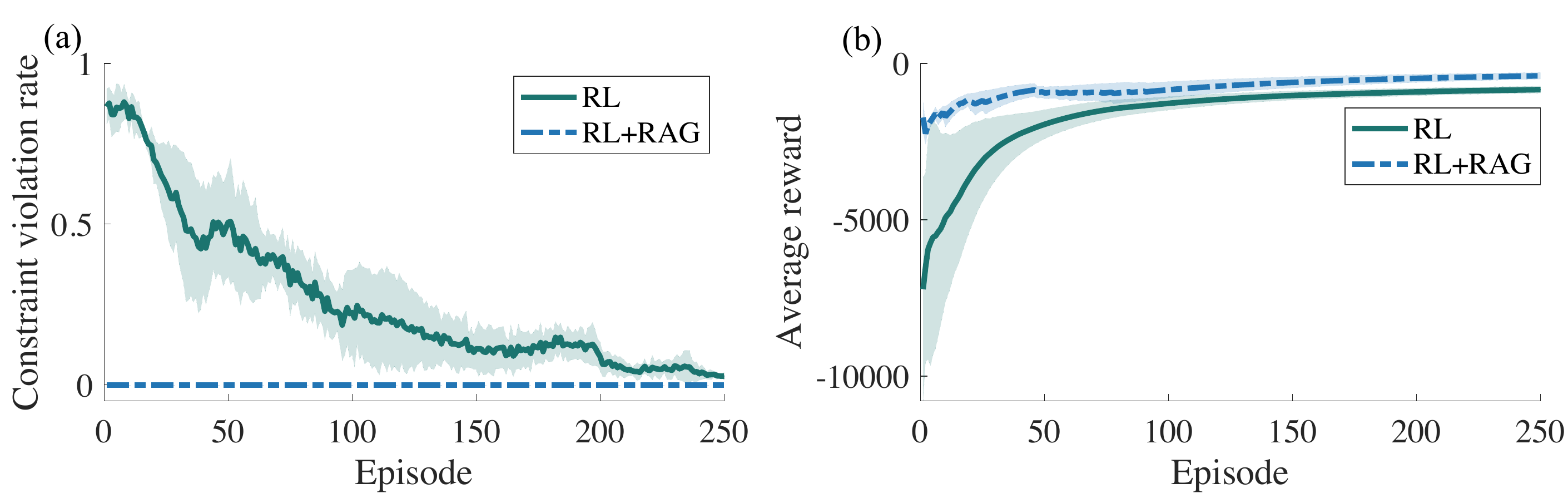}
      \caption{Training histories of conventional RL and safe RL. (a) Constraint violation rates of each episode. (b) Average reward values of each episode. Solid lines represent average values and shaded areas represent the standard deviation values over 20 experiments.}
      \label{fig:TrainResult}
\end{figure}

The training histories of conventional and safe RL algorithms are illustrated in Figure \ref{fig:TrainResult}. As the nominal feedback control policy tracks the headway time of 2.5s before training, which is not within the range of constraints in (\ref{equ:ACCSpec}), the constraint violation rate of conventional RL is high at the beginning of training. In contrast, no constraint violation is exhibited for safe RL during the entire training process. With RAG, the RL agent can learn the control policy safely without any constraint violation. Moreover, with RAG the headway time of ego vehicle is always within the range of $[1,2]$s, leading to a smaller reward variation and faster learning compared with conventional RL, as shown in Figure \ref{fig:TrainResult}b. 

\begin{figure}[thpb]
      \centering
      \includegraphics[scale=0.65]{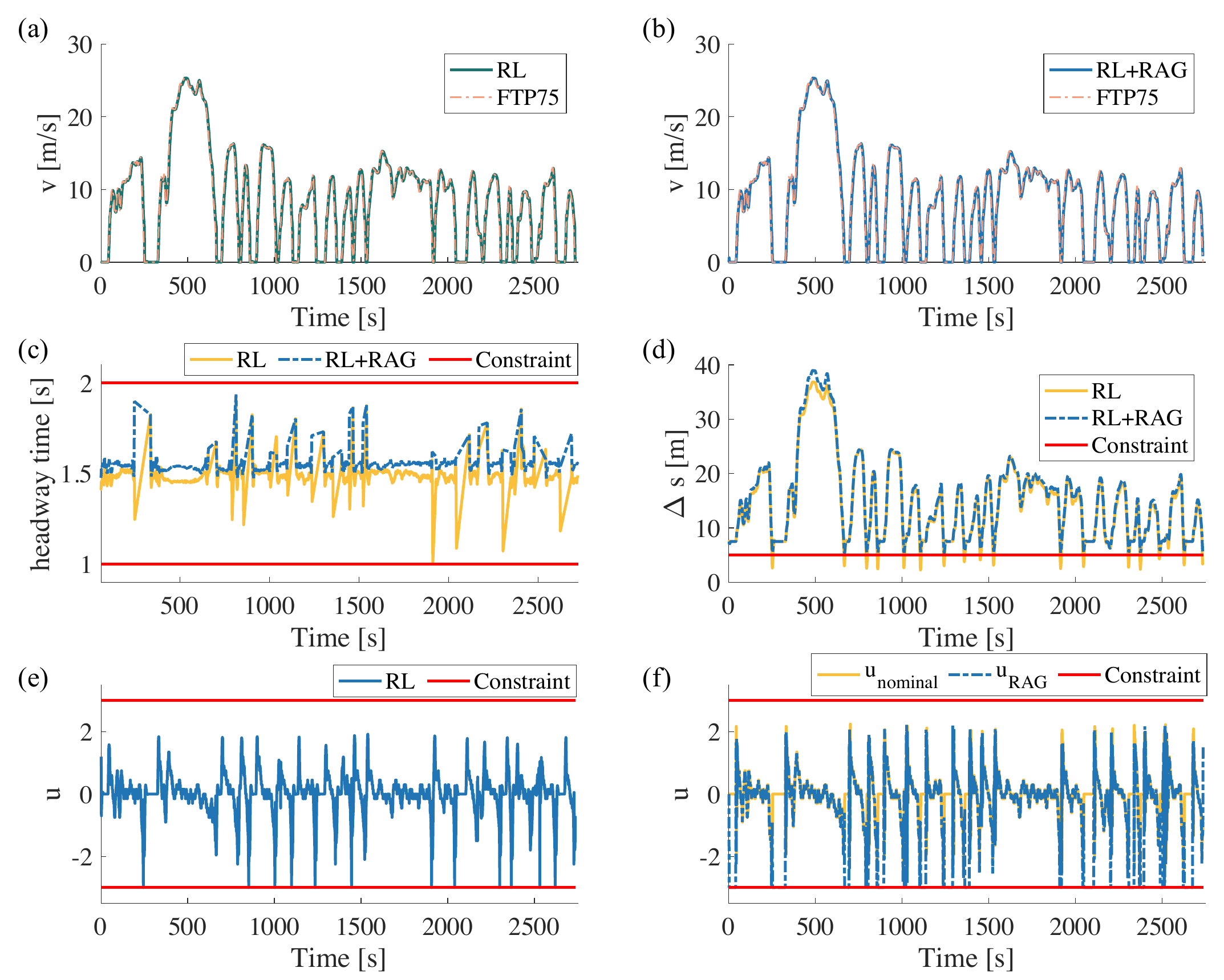}
      \caption{Validation results of conventional RL and safe RL. (a, b) Speed tracking performances. (c) headway time tracking performances (target value: 1.5s). (d) Relative distance between lead vehicle and ego vehicle. (e, f) Control inputs.}
      \label{fig:ValidateResult}
\end{figure}

The validation results of trained policies are shown in Figure \ref{fig:ValidateResult}. The policy trained with safe RL is implemented with RAG. As shown in Figure \ref{fig:ValidateResult}a, \ref{fig:ValidateResult}b and \ref{fig:ValidateResult}c, the velocity and headway time tracking performances of conventional and safe RL policies are both satisfactory. However, under the control of conventional RL policy, there are still occasional constraint violations in $\Delta s$, as shown in Figure \ref{fig:ValidateResult}d. In comparison, there is no constraint violation with the control of safe RL policy, which is attributed to the fact that the RAG monitors and modifies the control input to guarantee the constraint satisfaction as illustrated in Figure \ref{fig:ValidateResult}f. 

We remark that the fact that our safe RL algorithm based on the use of RAG guarantees no safety constraint violation during both the training and the operating phases yields that it can be used for onboard applications. For instance, it can be used to continuously improve the performance of a controller during its onboard operation.

\begin{figure}[thpb]
      \centering
      \includegraphics[scale=0.5]{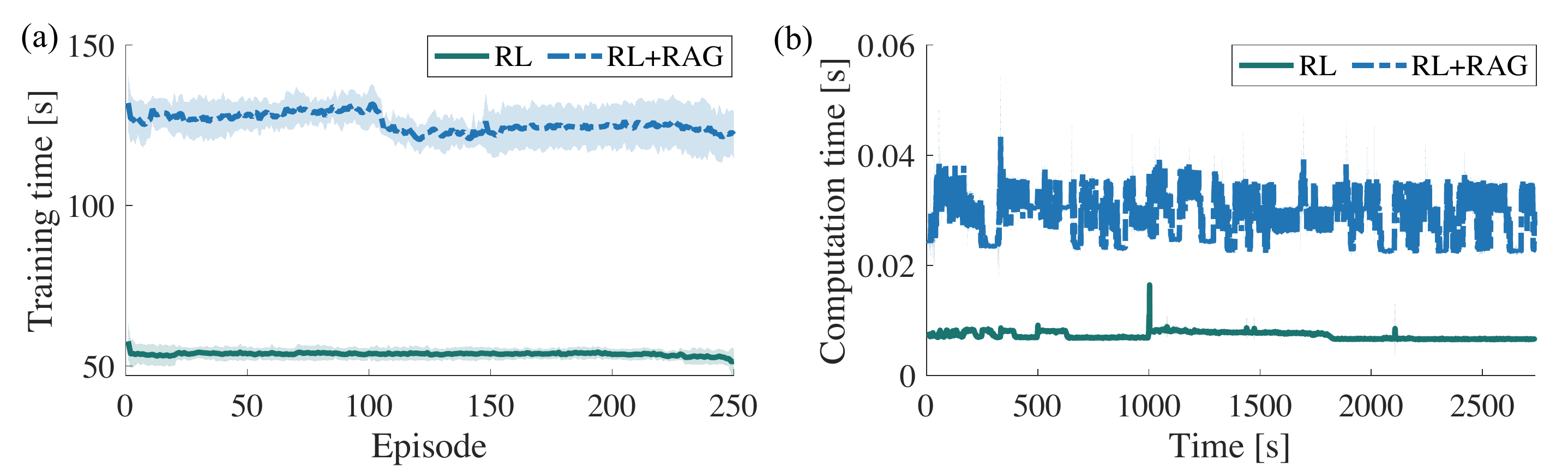}
      \caption{Comparison of computational times. (a) Training time of each episode. (b) Computational time of each step during validation. Solid lines represent average values and shaded areas represent the standard deviation values over 20 experiments.}
      \label{fig:ComputeTime}
\end{figure}

The computational time performances are shown in Figure \ref{fig:ComputeTime}. The simulations are performed on the Matlab R2019b platform using an Intel Xeon E5-1650 3.50 GHz PC with Windows 10 and 16.0 GB of RAM. As the optimization problem (\ref{equ:RAG}) needs to be solved at each time instant, safe RL takes around 130 s on average to complete an episode, which is longer compared with the one with conventional RL. Furthermore, we can observe that the average online solving time for (\ref{equ:RAG}) is around 30 ms, which is feasible for real-time control scenarios. 

\section{Conclusions and Future Work}\label{sec:6}
In this paper, we developed a safe RL framework based on RAG that integrates model-based safety supervision and model-free learning. We exploited the underlying dynamics and exclusion-zone requirement to construct a safety set for constraining learning exploration using set-theoretic techniques and online optimization. We applied the proposed safe RL framework to an Adaptive Cruise Control system and showed that we could conduct the online learning with no safety constraint violations. Future work will include the investigation into the computational complexity, scalability, and approaches to improving the exploration efficiency of the proposed safe RL framework.

\acks{This work is supported by Ford Motor Company. We also thank Mr. Xintao Yan for helpful discussions about implementing reinforcement learning algorithm in the ACC example.}

\bibliography{references}

\end{document}